\documentclass[10pt,twocolumn,letterpaper]{article}

\usepackage[T1]{fontenc}
\usepackage[letterpaper,margin=0.72in,top=0.73in,bottom=0.77in]{geometry}
\usepackage{amsmath,amssymb}
\usepackage{newtxtext,newtxmath}
\usepackage{microtype}
\usepackage{graphicx,booktabs,multirow,tabularx}
\usepackage[table]{xcolor}
\usepackage{float}
\usepackage[font=small,labelfont=bf,labelsep=period]{caption}
\usepackage[numbers,sort&compress]{natbib}
\usepackage{placeins}

\definecolor{cvprblue}{rgb}{0.21,0.49,0.74}
\definecolor{vasrResultHighlight}{HTML}{E4E2F3}
\newcommand{\vasrBest}[1]{\textcolor{red!72!black}{\textbf{#1}}}
\newcommand{\vasrDelta}[1]{\textcolor{black!75}{\scriptsize\ensuremath{#1}\,pp}}
\newcommand{\vasrRef}{{\scriptsize ref.}}
\newcommand{\vasrNA}{/}
\newcommand{\cmark}{\ensuremath{\checkmark}}
\newcommand{\xmark}{\ensuremath{\times}}

\newcommand*{\titlelead}{VASR: Vector-Aware Assembly Planning}
\newcommand*{\titletail}{with Equivalent Structures and Repair-Cost Feedback}
\newcommand*{\papertitle}{\titlelead\space\titletail}

\newcommand*{\paperauthornames}{Ruizhe Wang, Rui Liu, Hongliang Chen,
  Junyi Guo, Fangyu Wu, Yaran Chen}

\newcommand*{\paperauthors}{%
  Ruizhe Wang\textsuperscript{1}\enspace
  Rui Liu\textsuperscript{1,*}\enspace
  Hongliang Chen\textsuperscript{2,*}\enspace
  Junyi Guo\textsuperscript{1,\ensuremath{\dagger}}\enspace
  Fangyu Wu\textsuperscript{1,\ensuremath{\dagger}}\enspace
  Yaran Chen\textsuperscript{1,\ensuremath{\ddagger}}%
}
\newcommand*{\paperaffiliations}{%
  \textsuperscript{1}Xi'an Jiaotong-Liverpool University\quad
  \textsuperscript{2}The Hong Kong University of Science and Technology (Guangzhou)\\[2pt]
  {\small
    \textsuperscript{*}Co-second authors\quad
    \textsuperscript{\ensuremath{\dagger}}Co-third authors\quad
    \textsuperscript{\ensuremath{\ddagger}}Corresponding author:
    \href{mailto:Yaran.Chen@xjtlu.edu.cn}{\nolinkurl{Yaran.Chen@xjtlu.edu.cn}}%
  }%
}

\usepackage{hyperref}
\definecolor{linkink}{HTML}{214D72}
\hypersetup{
  pdftitle={\papertitle},
  pdfauthor={\paperauthornames},
  colorlinks=true,
  linkcolor=linkink,
  citecolor=linkink,
  urlcolor=linkink
}
\renewenvironment{abstract}
  {\noindent\textbf{Abstract.}\enspace\ignorespaces}
  {\par}

\begin{document}
\twocolumn[{%
  \begin{minipage}{\textwidth}
    \centering
    {\LARGE\bfseries\titlelead\\\titletail\par}
    \ifx\paperauthors\empty\else
      \vspace{0.8em}
      {\large\paperauthors\par}
      \ifx\paperaffiliations\empty\else
        \vspace{0.3em}
        {\normalsize\paperaffiliations\par}
      \fi
    \fi
    \vspace{1.15em}
    \begin{minipage}{0.92\textwidth}
      \begin{abstract}
Furniture assembly planning requires inferring subassembly hierarchies from individual parts and a target illustration. Existing raster-based methods do not explicitly exploit the geometric and structural information in SVG drawings. We introduce VASR, a framework that fuses features from a dedicated SVG encoder with visual representations from a pretrained vision–language model to predict assembly hierarchies. To avoid penalizing valid alternatives to a reference assembly tree, we propose Repair-Aware Hierarchical Policy Optimization (RHPO) with hierarchy-preserving equivalence rewards. RHPO further uses repair-cost feedback to favor decisions with lower downstream correction costs among candidates with comparable equivalence-aware rewards. VASR achieves the highest Equivalent-Hard F1 among the evaluated methods, outperforming GPT6-Astra by 17.69 percentage points on the IKEA-derived test set while maintaining strong performance on an independent furniture dataset.
\end{abstract}

    \end{minipage}
    \par\vspace{1.0em}
    \begin{minipage}{\textwidth}
\centering
\includegraphics[width=0.96\linewidth]{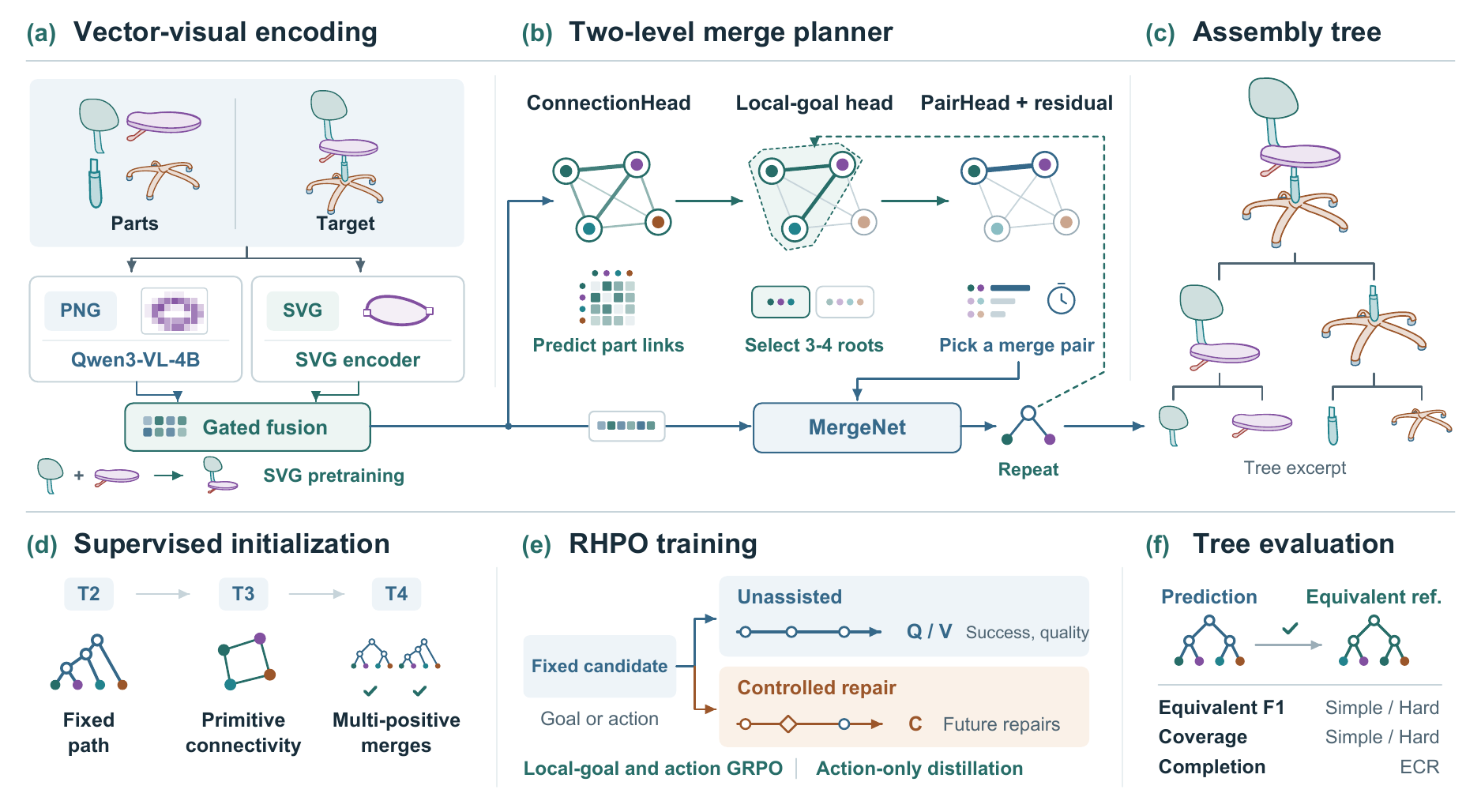}
\captionof{figure}{\textbf{Overview of VASR.}
(a) A learned gate fuses raster and SVG features from the part and target drawings.
(b) Predicted part connections guide local-goal selection. The action head scores candidate merge pairs conditioned on the current forest, goal, and timer. MergeNet encodes the selected pair as a parent node and returns it to the forest.
(c) Repeated pairwise merges produce a binary assembly tree.
(d) Supervised initialization progresses from a fixed assembly path to primitive connectivity and multi-positive merge supervision.
(e) RHPO evaluates fixed goals and actions through separate unassisted and controlled repair rollouts.
(f) Evaluation against equivalent references reports F1 and coverage under the Simple and Hard criteria, along with complete-tree accuracy.}
\label{fig:vasr-overview}
\end{minipage}

    \vspace{1.3em}
  \end{minipage}%
}]

\section{Introduction}

An illustration of an assembled chair reveals the relative positions of its legs, stretchers, and seat, but leaves open which parts should be assembled first. Furniture assembly planning requires organizing individual parts into intermediate subassemblies and determining how they combine into the target object. Understanding how components are progressively assembled from visual diagrams is a crucial step toward transforming human-readable assembly instructions into machine-actionable plans, thereby providing structured guidance for automated assembly and robot-assisted operations. An assembly tree captures this hierarchy and the dependencies among assembly steps~\cite{wang2022ikeamanual}. We study goal-conditioned tree prediction from unordered part illustrations and a final target illustration, without access to intermediate instruction pages at inference. Figure~\ref{fig:vasr-overview} summarizes our approach.

Prior work has used visual cues to infer assembly structure and actions.
Manual-PA matches 3D parts to step diagrams to infer their assembly
order~\cite{zhang2025manualpa}; TreeSBA predicts sequential brick
assemblies using a tree representation~\cite{guo2024treesba}; and
Manual2Skill parses manual pages with a vision--language model to build
hierarchical assembly graphs~\cite{tie2025manual2skill}. While raster
images convey shape and layout, they do not directly expose the drawing
primitives, coordinates, or path groupings available in native
SVG~\cite{cao2023svgformer}. These attributes can help distinguish similar
parts or identify where their shapes may fit together. They are cues,
not labels: an SVG group can be a drawing convention rather than a
physical subassembly. This distinction motivates learning a representation
of SVG structure rather than treating either the raw drawing syntax or
the SVG group hierarchy as an assembly plan~\cite{zou2024vgbench,xing2025llm4svg}.

Even with informative drawings, the finished object may admit several
assembly trees. Reordering the construction of two independent
subassemblies need not change the tree. By contrast, attaching the legs
to a chair seat before adding the backrest, or attaching the backrest to
the seat first, creates different intermediate subassemblies if both
choices are allowed. A single annotated tree can penalize a reasonable
alternative even when it preserves the required hierarchy. Early merge
errors also change the states encountered by subsequent
decisions~\cite{ross2011dagger}. Two candidates may therefore receive the
same terminal failure signal even when one requires a single correction
and the other requires repeated intervention.

We address these challenges with \textbf{VASR}, a vector-aware furniture
assembly planner. Its vector--visual representation combines the explicit
geometry of SVG drawings with pretrained visual features that capture
part appearance and the layout of the target. To recognize valid
alternatives, we define a hierarchy-preserving equivalence space that
allows interchangeable-part assignments and compatible refinements of
the reference hierarchy under annotated connectivity constraints. The
same equivalence rules inform multi-positive supervision, rollout
feedback, and evaluation.
An early merge error changes the states encountered by later decisions, allowing mistakes to compound over the remaining assembly process~\cite{ross2011dagger}. A branch that needs one local correction can receive the same terminal failure signal as a branch that needs repeated intervention. Starting from the supervised planner, we propose Repair-Aware Hierarchical Policy Optimization (RHPO), a variant of Group Relative Policy Optimization (GRPO)~\cite{shao2024deepseekmath}. RHPO compares candidates within the same decision context separately at each level, evaluating both local goals and merge actions by their consequences for the entire assembly.

VASR encodes each part and target drawing with an SVG encoder and a
pretrained visual encoder, then combines the two representations through
a learned gate. The planner predicts part connectivity and uses it to
score merges. A local-goal policy selects a short-horizon subassembly,
and an action policy repeatedly merges pairs of parts or existing
subassemblies to produce a complete tree. Multi-positive supervised
initialization teaches the planner to recognize more than one valid
merge in a state. RHPO then uses a frozen policy snapshot to compare
candidate goals and actions through separate unassisted and controlled
repair rollouts. Reference trees and annotated connections support
training-time supervision and verification. At test time, the planner
receives only the part and target drawings and uses its own predicted
connections, without intermediate instruction pages, reference trees,
annotated connections, or repair.

\textbf{Our contributions are threefold:}
\begin{itemize}
  \item We introduce VASR, which fuses native SVG geometry with
        pretrained visual features to predict subassembly hierarchies
        from unordered part drawings and an assembled-target drawing.
  \item We define hierarchy-preserving assembly equivalence for
        supervision and evaluation, and propose RHPO to use verified
        repair cost to distinguish candidate decisions with comparable
        unassisted outcomes.
  \item On the IKEA-derived and independently collected furniture test
        sets, VASR achieves the highest Equivalent-Hard F1 among the
        compared methods, exceeding GPT6.0-Astra by 17.69 and 14.55
        percentage points, respectively.
\end{itemize}

\section{Related Work}

\paragraph{Assembly Structure Prediction}

Existing work has explored assembly structure prediction under different forms of geometric and instructional supervision. IKEA-Manual establishes an assembly-tree benchmark from decomposed 3D parts and human-authored manuals \citep{wang2022ikeamanual}. Image- and diagram-guided approaches use target images or instructional illustrations to align parts and estimate their poses \citep{li2020singleimage,zhang2025manualpa}. Other methods represent sequential assembly as a tree or learn discrete part and connection-site decisions from geometric supervision, as demonstrated by TreeSBA and AssembleRL \citep{guo2024treesba,aslan2022assemblerl}. More recent work extracts hierarchical assembly graphs from step diagrams, explicitly models connector attributes, and recovers assembly trees from manual pages, including Manual2Skill and Manual2Skill++ \citep{tie2025manual2skill,tie2026manual2skillplusplus,qi2026assemstate}. VASR predicts the hierarchy from part and final-target illustrations without intermediate instruction pages, annotated connectivity graphs, or 3D pose estimation in inference.

\paragraph{Vector Representations and Visual Fusion}

Prior work on vector graphics has explored hierarchical representations of shapes and drawing commands, as well as direct modeling of continuous SVG parameters, as in DeepSVG and SVGformer \citep{carlier2020deepsvg,cao2023svgformer}. VGBench provides an evaluation benchmark for vector-graphics understanding and generation \citep{zou2024vgbench}. Recent approaches further introduce learnable semantic tokens or align SVG sequences with other modalities, including raster images, as demonstrated by LLM4SVG and InternSVG \citep{xing2025llm4svg,wang2026internsvg}. Related work also connects image features with drawing primitives for symbol spotting or combines SVG with pretrained image features for reconstruction and generation, as explored by CADTransformer and SVGFusion \citep{fan2022cadtransformer,xing2026svgfusion}. VASR uses vector and visual features for a different output: the hierarchy of depicted furniture parts and subassemblies. Its encoder learns task-specific correspondences rather than treating SVG path groups as assembly annotations.

\paragraph{Supervision over Alternative Structures}

Supervision becomes more complex when multiple action sequences or merge choices can lead to the same reference structure. Manual2Skill accounts for interchangeable components through permutations of equivalent identities in its pose objective \citep{tie2025manual2skill}. In imitation learning, dynamic oracles support alternative derivations of a reference structure, while DAgger acquires labels at states visited by the policy \citep{goldberg2013oracles,ross2011dagger}. AggreVaTe and LOLS further incorporate future costs or candidate continuations to construct cost-sensitive targets \citep{ross2014aggrevate,chang2015lols}. Our setting requires supervising several merges from the same partial assembly while retaining the reference hierarchy, rather than selecting one manual sequence or rewarding arbitrary connected trees.

\paragraph{Local Goals and Learning from Corrections}

Temporal abstraction provides a framework for decisions that span multiple actions \citep{sutton1999temporal}. Our local goals condition individual merges, but both decision levels are judged by full assembly outcomes. Unlike a hard action mask, a local goal biases scores without eliminating out-of-goal merges. Tree-GRPO and TreeRPO use branched rollouts to obtain intermediate feedback from downstream returns, while Rewarding Progress estimates the change in the probability of success under a continuation policy \citep{ji2026treegrpo,yang2025treerpo,setlur2025rewarding}. These methods motivate candidate-level evaluation, but more detailed return propagation need not distinguish branches whose sampled outcomes are nearly identical.

Prior work has incorporated human interventions into policy learning by penalizing intervention costs or learning from intervention events, as explored by HACO and RLIF \citep{li2022haco,luo2024rlif}. More recent approaches incorporate repaired observations, counterfactual replay, or sparse corrections into policy optimization. Find, Fix, Reason uses repaired observations and an optional fixed intervention penalty in GRPO, while PivotOPD uses counterfactual replay to distill corrective and recovery actions \citep{huang2026findfixreason,he2026pivotopd}. MInTRL applies sparse corrections and then returns control to the rollout policy \citep{chen2026mintrl}. RHPO retains the candidate decision and repairs only its subsequent errors. The resulting repair burden is kept separate from unassisted success and used as a preference only when comparable quality and a cost difference are supported. Unlike standard policy distillation from a teacher distribution \citep{rusu2016distillation}, our action target is formed by reallocating probability from confirmed incorrect actions within the frozen policy's own distribution.

\section{Method}

\subsection{Task and Equivalent Assembly Trees}

Given \(n\) parts, VASR takes the raster and native SVG drawings of
each part and of the target object:
\begin{equation}
x_0 = \bigl(\{(I_i,S_i)\}_{i=1}^{n}, I_\star, S_\star\bigr).
\label{eq:input}
\end{equation}
The output is a rooted tree whose leaves are the \(n\) primitive parts,
each appearing once. Each internal node denotes a subassembly, identified
by its descendant leaf set (clade). Starting from singleton parts, the
decoder merges two current roots at a time, yielding \(n-1\) merges.
Reordering independent merges leaves the tree unchanged.

An equivalent tree preserves all reference clades under a single
permutation of geometrically interchangeable parts and may add compatible
nested clades. Each non-leaf clade must induce a connected subgraph of
the annotated part graph \(G^*\). The equivalent space \(\mathcal E(x)\)
for instance \(x\) allows
multiway nodes, while our decoder targets
\begin{equation}
\mathcal E_{\mathrm{bin}}(x)
= \mathcal E(x) \cap \{\text{binary assembly trees}\}.
\label{eq:equivalent-binary-trees}
\end{equation}
The reference tree and annotated graph are used for training and
evaluation, not inference. Graph connectivity constrains equivalence
but does not guarantee physical executability.

\subsection{Task-Oriented SVG Representation}

\paragraph{SVG pretraining and composition.}
We extract part and target SVGs from assembly illustrations.
Pretraining aligns independently pose-augmented part drawings with
canonical components and learns subassembly representations from
same-source equivalence and membership supervision.
Appendix~\ref{sec:method-supplement} details the pretraining objectives.
To compose an assembly state \(s\), we embed each child representation
\(z_i\) jointly with its shared-canvas descriptor \(g_i\) as
\(h_i=\operatorname{MLP}([z_i,g_i])\) and form pair descriptors \(p_{ij}\).
For \(m\) children, the layout-aware composer adds a residual to their
normalized mean:
\begin{equation}
\begin{aligned}
z_{\mathrm{base}}
  &= \operatorname{norm}\!\left(\frac{1}{m}\sum_i z_i\right),\\
\hat z_s
  &= \operatorname{norm}\!\bigl(z_{\mathrm{base}}
      + \alpha\Delta_\omega[
      z_{\mathrm{base}}, \operatorname{mean}_i h_i,\\
  &\hspace{6em} \operatorname{mean}_{i<j} p_{ij},
      \operatorname{card}(s)]\bigr).
\end{aligned}
\label{eq:layout-composer}
\end{equation}
These descriptors are internal to SVG composition, not a separate
downstream input.

\paragraph{Gated fusion and recursive merging.}
We pool the final image-token states of
Qwen3-VL-4B-Instruct~\cite{bai2025qwen3vl} to obtain \(r_i\), then fuse
it with the SVG embedding \(z(S_i)\) through a learned gate:
\begin{equation}
h_i = P_v(r_i) + \sigma(g_s)\sqrt{d}\,z(S_i)
\label{eq:fusion}
\end{equation}
The same fusion yields the target feature \(h_x\). For a canonically
ordered pair of frontier nodes \(A,B\), PairHead scores the pair
features and MergeNet constructs the parent:
\begin{equation}
\begin{aligned}
\psi(A,B,x) &= [h_A,h_B,|h_A-h_B|,h_A\odot h_B,h_x],\\
h_{A\cup B} &= \operatorname{MergeNet}_\theta(\psi(A,B,x)).
\end{aligned}
\label{eq:recursive-merge}
\end{equation}
The merge replaces the two roots with their parent while retaining
their subtrees and primitive memberships.

\subsection{Supervised Structural Initialization}

\subsubsection{T2 Fixed-Path Supervision}

We initialize the planner with the Qwen3-VL adapter and the pretrained SVG encoder. T2 uses teacher forcing along a deterministic binary refinement. For reference action \(a_{t}^{*}\) at state \(s_{t}\), PairHead is normalized over all unordered pairs of current roots:

\begin{equation}
\begin{aligned}
p_\theta(a\mid s_t,x_0) &= \operatorname{softmax}_{a\in\mathcal{A}(s_t)}
                             \ell_\theta(a), \\
L_{\mathrm{fixed}} &= -\frac{1}{n-1}\sum_t
                        \log p_\theta(a_t^\star\mid s_t,x_0).
\end{aligned}
\label{eq:fixed-path}
\end{equation}

After following the reference trajectory, we align the final root with the target representation using a stop-gradient target:

\begin{equation}
\begin{aligned}
L_{\mathrm{final}} &= 1-\cos(h_{\mathrm{root}},\operatorname{sg}(h_x))
\end{aligned}
\label{eq:t2}
\end{equation}

\subsubsection{T3 Learning Primitive Connectivity}

ConnectionHead maps the fused features of each original primitive pair and the target to one logit, \(z_{ij}\). We evaluate each unordered pair once and apply a sigmoid to obtain \(p_{ij}\), the predicted direct connection score. The resulting matrix is symmetric by construction and has a zero diagonal. It contains \(n(n-1)/2\) distinct predictions. These are scores for all possible pairs, not guaranteed true edges or transitive reachability.

\begin{equation}
p_{ij}=\sigma(z_{ij})=p_{ji}
\label{eq:predicted-connectivity}
\end{equation}

We supervise \(p_{ij}\) using \(y_{ij}\), the binary edge label of \(G^*\). Let \(p_{ij}(y)\) be \(p_{ij}\) for a positive label and \(1-p_{ij}\) for a negative label. The focal objective uses \(\gamma=2\) and clips the positive-class weight to \([1,10]\):

\begin{equation}
\begin{aligned}
L_{\mathrm{conn}} &= \frac{1}{|\mathcal{P}|}\sum_{i<j}
    (1-p_{ij}^{(y)})^\gamma\operatorname{BCE}_{w_+}(p_{ij},y_{ij}), \\
\gamma &= 2, \qquad
w_+ = \operatorname{clip}\!\left(\frac{N_-}{\max(N_+,1)},1,10\right).
\end{aligned}
\label{eq:connectivity-loss}
\end{equation}

T3 aggregates the predicted primitive scores across the two subassemblies by their arithmetic mean. This is the existing T4 merge-score bias.

\begin{equation}
\begin{aligned}
\bar p_{AB} &= \frac{1}{|A||B|}\sum_{i\in A,\,j\in B}p_{ij}, \\
\ell_\theta^{\mathrm{T3}}(A,B) &= f_{\mathrm{pair}}(\psi)
                                +\operatorname{softplus}(\alpha_0)\bar p_{AB}
\end{aligned}
\label{eq:t3}
\end{equation}

\begin{figure*}[t]
    \centering
    \includegraphics[width=\linewidth]{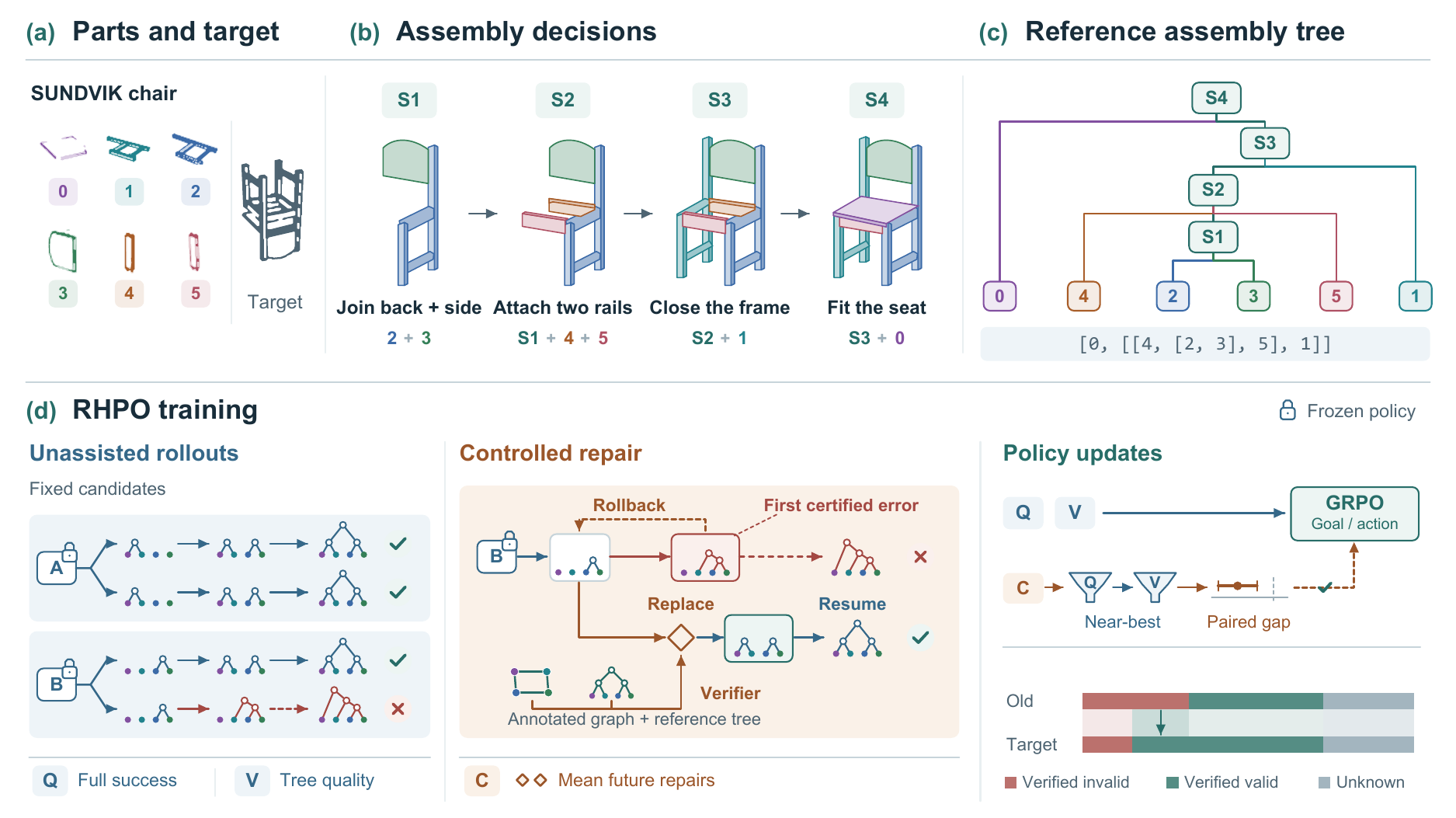}
    \caption{\textbf{Assembly architecture and RHPO training} (a–c) The indexed parts and target drawing illustrate four assembly steps. Each resulting subassembly maps to an internal node of the reference tree, also shown in nested JSON. (d) For a fixed goal or action, unassisted rollouts estimate full success (Q) and tree quality (V). Separate repair rollouts estimate the mean number of subsequent repairs (C). At the first certified downstream error, the verifier rolls back, replaces the erroneous merge using the annotated graph and reference tree, and resumes the frozen policy. GRPO uses Q and V at both decision levels and favors lower repair cost among eligible candidates with near-best quality and a certified paired cost gap. Action-only distillation transfers probability mass from verified invalid actions to verified valid actions while preserving the mass on unknown actions.}
    \label{fig:vasr-rhpo}
\end{figure*}

\subsubsection{T4 Dynamic Multi-Positive Supervision}

For an assembly state \(s_{t}\), the training oracle constructs a set of admissible next merges. We replace single-target cross-entropy with the negative log of the set's total probability mass, retaining the connection and final-state objectives:

\begin{equation}
\begin{aligned}
L_{\mathrm{set}} &= -\frac{1}{n-1}\sum_t
                    \log\sum_{a\in\mathcal{A}_{\mathrm{T4}}^+(s_t)}
                        p_\theta(a\mid s_t,x_0)
\end{aligned}
\label{eq:t4}
\end{equation}

At each step, we sample the next merge from the oracle set using the model's logits at temperature 1. The oracle checks geometric-equivalence multisets and compatibility with the reference clades. It prefers cross-edge-supported candidates and falls back to hierarchy-compatible candidates when none are available. If the local equivalence check leads to a trajectory that cannot be completed, we roll back and regenerate it with the original reference part labels. This practical T4 oracle is distinct from the explicit RHPO verification protocol below.

\subsection{Two-Level Control on the Predicted Graph}

RHPO adds a local-goal head and an action residual head to T4, with the shared planner and ConnectionHead frozen. The local head selects a short-horizon subassembly from subsets of three or four current roots that cover only part of the forest. The proposal affinity for two roots is the strongest predicted primitive edge across their memberships. A maximum-spanning-tree bottleneck ranks subsets by the weakest edge in their maximum spanning tree. The decoder keeps up to eight proposals and uses the null goal when the proposal set is empty.

The selected goal persists as a set of primitive parts until completion, a merge that mixes in-goal and out-of-goal parts, a three-action horizon, or termination. The action head conditions on the full forest, goal, and timer. It adds a bounded goal-progress term and a learned residual to the T4 merge logit. Subtracting the initial residual and bounding the goal term by half the leading T4 score gap preserves T4's initial greedy choice under matched decoding conditions. Every unordered pair remains in the action support. The maximum affinity ranks goal proposals, and the T4 mean edge score continues to bias merges.

The predicted graph \(\hat G\) supplies connectivity to the policy in both training rollouts and inference. The annotated graph \(G^*\) supplies edge labels and supports the training verifier and repairs.

\subsection{Frozen Rollouts and Repair Preferences}

\paragraph{Evaluate one candidate at a time.}
For each decision level \(\ell\), we force a candidate \(y\) in a full context \(c\) and resume the same frozen two-level controller. A forced goal remains active for its normal duration, and a forced action takes effect once. Write \(I_y\) for this intervention on the initial choice. Unassisted rollouts estimate the probability of a complete equivalent tree and expected terminal structural quality.
\begin{equation}
  \begin{aligned}
  Q_r^\ell(c,y) &= \Pr(Y(T,x)=1\mid I_y), \\
  V_r^\ell(c,y) &= \mathbb{E}[F(T,x)\mid I_y].
  \end{aligned}
  \label{eq:unassisted-main}
\end{equation}
Here \(Y\) indicates equivalence and \(F\in[0,1]\) is the structural score. Both levels are evaluated on the entire assembly.

\paragraph{Measure downstream repair.}
The repair rule \(\kappa\) runs separately. On failure, the training verifier rolls back to immediately before the first certified erroneous merge after the candidate, replaces it with a uniformly sampled verified valid merge, updates the goal state, and returns control to the frozen controller. The candidate stays fixed. We estimate
\begin{equation}
  C_r^{\ell,\kappa}(c,y)=\mathbb{E}[N_{\mathrm{future}}\mid I_y,\kappa],
  \label{eq:repair-cost-main}
\end{equation}
where \(N_{\mathrm{future}}\) counts subsequent interventions. Inconclusive or exhausted trials require a valid cost bound for certification. Repair cost describes the continuation behavior of the frozen policy under \(\kappa\).

\paragraph{Screen quality before comparing costs.}
Using simultaneous confidence bounds, first retain candidates whose lower bound on \(Q\) lies within \(\varepsilon_Q\) of the best upper bound on \(Q\) among queried candidates. Within that set, apply the same test to \(V\) with tolerance \(\varepsilon_F\). Actions must also be verified valid, and local goals must start from an extendable forest. Independent paired repair trials compare each candidate with a quality baseline selected by a fixed rule. With \(n_{\mathrm{pair}}\) independent pairs, mean cost difference \(\bar d\), remaining horizon \(H_c\), and one-sided error allocation \(\alpha'\), certify lower cost when
\begin{equation}
  \bar d + H_c\sqrt{\frac{2\log(1/\alpha')}{n_{\mathrm{pair}}}}
  < -\delta_C.
  \label{eq:cost-certification-main}
\end{equation}
If fewer than two distinct candidates qualify or the cost difference cannot be certified, the cost preference is omitted.

\subsection{Optimization and Conservative Updates}

\paragraph{Group-relative policy updates.}
Once the candidate estimates are fixed, draw fresh update groups at each level from the frozen old policy. Each action group shares the full context, including the goal and timer. Repeated candidates retain their sampling frequency. We standardize \(R_0=Q+\omega_F V\) within each group to obtain the quality advantage \(A_0\), with a floor on the standard deviation. For the baseline and certified lower-cost candidates, we mean-center \(-C/\max(1,H_c)\) within their subgroup, where \(H_c\) is the remaining binary horizon. The cost advantage \(A_C\) is zero outside the subgroup and whenever fewer than two distinct eligible candidates occur. Using the clipped PPO surrogate~\cite{schulman2017ppo} and group-relative advantages~\cite{shao2024deepseekmath}, the loss at either level is
\begin{equation}
  \mathcal{L}_{\mathrm{RL}}^\ell =
  \mathcal{L}_{\mathrm{clip}}^\ell(A_0)+
  s_{C,\mathrm{RL}}\beta_C^\ell\mathcal{L}_{\mathrm{clip}}^\ell(A_C).
  \label{eq:policy-loss-main}
\end{equation}
The estimates, gates, and advantages are detached. 

\paragraph{Conservative action distillation.}
Partition the old action distribution \(p\) into verified positives \(P\), verified negatives \(N\), and unknowns \(U\). The target transfers mass \(\Delta=\eta p(N)\) from \(N\) to \(P\).
\begin{equation}
  t(a)=
  \begin{cases}
  p(a)+\Delta[(1-\lambda)p(a\mid P)+\lambda w(a)], & a\in P,\\
  (1-\eta)p(a), & a\in N,\\
  p(a), & a\in U.
  \end{cases}
  \label{eq:distill-main}
\end{equation}
Skip contexts with zero positive mass. The distribution \(w\) is supported on verified positives. With the distillation cost switch enabled, it favors certified lower-cost actions when available. It can also use reliable unassisted-success evidence. Without either source of evidence, \(\lambda=0\). The cost switch for distillation is independent of \(s_{C,\mathrm{RL}}\). A KL loss fits the detached target on visited and replay states. The target preserves unknown probabilities and never reduces a verified positive's mass. 
\paragraph{Update checks and inference.}
We alternate action and local rounds, with replay protection and a KL penalty to the fixed initialization. At each round, we refresh the old policy. Before accepting an update, we audit the change in unassisted success with the continuation fixed, check valid-action mass and decision margins on action replay anchors, and require the fixed-batch surrogate loss to decrease or stay constant. The checks apply to the audited contexts. In inference the decoder predicts connectivity and builds the tree from the input illustrations using the two learned heads. The verifier and repair mechanism are confined to training. 

\section{Experiments}
\subsection{Experimental Setup}
\paragraph{Datasets}
The IKEA dataset~\cite{wang2022ikeamanual} contains 102 sources and 1,020 pose instances, with all 10 poses of each source kept in the same split. We report averages across five cross-validation folds. We also evaluate VASR on an independently collected dataset of 100 furniture objects to assess generalization.
\paragraph{Baselines}
We compare VASR with task-specific assembly planners from IKEA-Manual and Manual2Skill~\cite{wang2022ikeamanual,tie2025manual2skill} and six general-purpose vision--language models: Qwen3-VL-32B, GLM-5.3-Flash, Qwen3.8-27B, GPT6-Sol, GPT6-Astra, and DeepSeek-V4.1-Flash~\cite{bai2025qwen3vl,zai2026glm53flash,qwen2026qwen3827b,openai2026gpt6sol,openai2026gpt6astra,deepseekai2026deepseekv41flash}.

\begin{table*}[t]
  \centering
  \small
  \setlength{\tabcolsep}{2.6pt}
  \renewcommand{\arraystretch}{1.05}
  \definecolor{vasrResultHighlight}{HTML}{E4E2F3}

  \sbox0{%
    \begin{tabular}{@{}clrr@{\,}lrrr@{\hspace{8pt}}rr@{\,}lrrr}
      \toprule
      & \multirow{2}{*}{\textbf{Method}}
      & \multicolumn{6}{c}{\textbf{IKEA-Derived Test Set}}
      & \multicolumn{6}{c}{\textbf{Independent Furniture Test Set}} \\
      \cmidrule(lr){3-8}\cmidrule(l){9-14}
      & & \multicolumn{1}{c}{Simple} & \multicolumn{2}{c}{Hard / \(\Delta\)}
        & \multicolumn{1}{c}{ECS} & \multicolumn{1}{c}{ECH} & \multicolumn{1}{c}{ECR}
        & \multicolumn{1}{c}{Simple} & \multicolumn{2}{c}{Hard / \(\Delta\)}
        & \multicolumn{1}{c}{ECS} & \multicolumn{1}{c}{ECH} & \multicolumn{1}{c}{ECR} \\
      \midrule
      & Qwen3-VL-32B~\cite{bai2025qwen3vl}
        & 28.69\% & 12.32\% & & 28.00\% & 11.90\% &  3.75\%
        & 26.73\% &  9.35\% & & 24.14\% & 10.03\% &  3.64\% \\
      & GLM-5.3-Flash~\cite{zai2026glm53flash}
        & 37.22\% & 17.11\% & & 35.98\% & 22.07\% & 13.13\%
        & 38.14\% & 18.72\% & & 37.72\% & 23.50\% & 15.63\% \\
      & Qwen3.8-27B~\cite{qwen2026qwen3827b}
        & 47.47\% & 25.58\% & & 41.94\% & 33.90\% & 20.00\%
        & 47.14\% & 26.31\% & & 42.45\% & 34.10\% & 22.10\% \\
      & DeepSeek-V4.1-Flash~\cite{deepseekai2026deepseekv41flash}
        & 47.40\% & 22.19\% & & 42.69\% & 27.42\% &  9.68\%
        & 46.51\% & 21.49\% & & 40.87\% & 26.30\% &  9.14\% \\
      & GPT6.0-Sol~\cite{openai2026gpt6sol}
        & 54.99\% & 30.88\% & & 47.55\% & 42.97\% & 22.50\%
        & 56.25\% & 32.16\% & & 50.31\% & 44.82\% & 24.72\% \\
      \multirow{-6}{*}{\rotatebox[origin=c]{90}{\scriptsize\bfseries Generalist VLMs}}
      & GPT6.0-Astra~\cite{openai2026gpt6astra}
        & 61.04\% & 37.12\% & \vasrDelta{-17.69} & 52.64\% & 50.66\% & 25.00\%
        & \vasrBest{68.71} & 39.15\% & \vasrDelta{-14.55} & 57.46\% & 53.10\% & 27.72\% \\
      \midrule
      & Manual2skill~\cite{tie2025manual2skill}
        & 57.30\% & 32.40\% & & 60.92\% & 34.10\% & 22.76\%
        & 50.71\% & 29.89\% & & 50.92\% & 31.13\% & 20.37\% \\
      & IKEA-SingleStep~\cite{wang2022ikeamanual}
        & 48.64\% & 11.42\% & & 45.14\% & 11.42\% &  4.31\%
        & \vasrNA & \vasrNA & & \vasrNA & \vasrNA & \vasrNA \\
      \multirow{-3}{*}{\rotatebox[origin=c]{90}{\tiny\bfseries Specific models}}
      & IKEA-GeoCluster~\cite{wang2022ikeamanual}
        & 43.53\% & 16.93\% & & 44.12\% & 18.93\% & 8.08\%
        & \vasrNA & \vasrNA & & \vasrNA & \vasrNA & \vasrNA \\
      \midrule
      & VASR-S
        & 50.43\% & 33.75\% & \vasrDelta{-21.06} & 56.06\% & 48.91\% & 24.49\%
        & 48.03\% & 32.82\% & \vasrDelta{-20.88} & 54.55\% & 46.15\% & 22.16\% \\
      & \enspace \(+\) GRPO
        & 51.16\% & 31.68\% & \vasrDelta{-23.13} & 57.73\% & 46.42\% & 23.52\%
        & 48.10\% & 29.18\% & \vasrDelta{-24.52} & 55.39\% & 44.58\% & 21.89\% \\
      & \enspace \(+\) TreeRPO
        & 58.15\% & 34.86\% & \vasrDelta{-19.95} & 55.60\% & 49.83\% & 25.56\%
        & 58.81\% & 33.14\% & \vasrDelta{-20.56} & 56.13\% & 50.07\% & 26.20\% \\
      & \enspace \(+\) Tree-GRPO
        & 60.32\% & 38.85\% & \vasrDelta{-15.96} & 63.82\% & 52.16\% & 28.70\%
        & 59.08\% & 36.25\% & \vasrDelta{-17.45} & 60.26\% & 51.18\% & 27.42\% \\
      \rowcolor{vasrResultHighlight}
      \cellcolor{white}\multirow{-5}{*}{\rotatebox[origin=c]{90}{\scriptsize\bfseries VASR family}}
      & \textbf{VASR (full)}
        & \vasrBest{67.73} & \vasrBest{54.81} & \vasrRef & \vasrBest{77.35} & \vasrBest{62.43} & \vasrBest{33.17}
        & 66.31\% & \vasrBest{53.70} & \vasrRef & \vasrBest{74.33} & \vasrBest{60.87} & \vasrBest{31.74} \\
      \bottomrule
    \end{tabular}%
  }
  \ifdim\wd0>\textwidth
    \resizebox{\textwidth}{!}{\usebox{0}}%
  \else
    \usebox{0}%
  \fi
  \caption{\textbf{Assembly planning on two datasets.}
  Scores are percentages. The best measured score for each metric on
  each dataset is shown in \textcolor{red!72!black}{\textbf{red}}.
  A slash (/) indicates an unavailable measurement.
  IKEA-SingleStep and IKEA-GeoCluster are the two IKEA-Manual
  planning baselines~\cite{wang2022ikeamanual};
  Manual2Skill is from Tie et al.~\cite{tie2025manual2skill}.
  Simple and Hard denote Equivalent-Simple and Equivalent-Hard F1.
  For GPT6.0-Astra and the VASR variants, \(\Delta\) is the Hard F1
  difference from full VASR on the same dataset (method minus VASR),
  in percentage points (pp). VASR-S is supervised only;
  \(+\) denotes further reinforcement learning.}
  \label{tab:vasr-main-results}
\end{table*}

\paragraph{Evaluation Metrics}
We extend the \textit{Simple} and \textit{Hard} metrics of IKEA-Manual~\cite{wang2022ikeamanual} to account for equivalent trees. We report Equivalent-Simple and Equivalent-Hard F1, the equivalent completion rate (ECR), and two equivalent coverage scores (ECS and ECH).
All node counts include the root and exclude primitive-part leaves.

Equivalent-Simple matches nodes containing the same set of primitive
parts. Equivalent-Hard additionally requires identical unordered
partitions into immediate children, including singleton children.
For each criterion $c\in\{\mathrm{S},\mathrm{H}\}$, we independently
select a reference maximizing the instance-level F1:
\begin{equation}
T_c^* \in
\underset{T'\in\mathcal{E}(T,G)}{\arg\max}
\frac{2M_c(\widehat{T},T')}
     {N(\widehat{T})+N(T')},
\end{equation}
where $\widehat{T}$ is the predicted tree, $N(\cdot)$ counts non-leaf
nodes, and $M_c(\cdot,\cdot)$ counts matched nodes under criterion $c$.
Precision, recall, and F1 are then computed against the same selected
reference:
\begin{equation}
\begin{gathered}
P_c = \frac{M_c(\widehat{T}, T_c^*)}{N(\widehat{T})},\qquad
R_c = \frac{M_c(\widehat{T}, T_c^*)}{N(T_c^*)}, \\
F_{1,c} = \frac{2M_c(\widehat{T}, T_c^*)}
                 {N(\widehat{T}) + N(T_c^*)}.
\end{gathered}
\end{equation}

ECR measures the proportion of predictions that themselves constitute
admissible equivalent trees:
\begin{equation}
\mathrm{ECR}
=
\frac{1}{K}\sum_{i=1}^{K}
\mathbf{1}\!\left[
\widehat{T}_i\in\mathcal{E}(T_i,G_i)
\right],
\end{equation}
where $K$ is the number of evaluated instances. This is equivalent
to the fraction achieving instance-level Equivalent-Hard F1 of one.

We retain ECS and ECH as historical node-coverage diagnostics:
\begin{equation}
\begin{aligned}
\mathrm{ECS}
&= \frac{1}{K}\sum_{i=1}^{K}
\frac{M_{\mathrm{legacy},\mathrm{S},i}}{N(T_i)}, \\
\mathrm{ECH}
&= \frac{1}{K}\sum_{i=1}^{K}
\frac{M_{\mathrm{legacy},\mathrm{H},i}}{N(T_i)}.
\end{aligned}
\end{equation}
ECS matches nodes by their counts of geometrically equivalent part
classes. ECH uses maximum one-to-one node matching with the same
part-set or equivalence-class signatures.

We report instance-macro averages of precision, recall, and F1,
averaging each independently. Invalid or incomplete predictions
receive zero scores. Instances with no admissible equivalent reference
receive zero Equivalent-Simple, Equivalent-Hard, and ECR scores.

\subsection{Main Results}

Table~\ref{tab:vasr-main-results} reports results on the IKEA-derived
test set and an independently collected furniture test set. We compare
VASR with three task-specific furniture assembly planning methods and models, six generalist
vision--language models (VLMs), and three reinforcement learning (RL)
variants initialized from the supervised-only VASR-S. 

\paragraph{Comparison with task-specific assembly baselines.}
The IKEA comparison uses information from the assembly manuals for
all three task-specific baselines, but their inputs differ.
Manual2Skill builds an assembly graph from step
illustrations~\cite{tie2025manual2skill}. The IKEA-Manual heuristics
instead operate on a set of parts: SingleStep places them under one
root, and GeoCluster groups them using features from 3D part point
clouds~\cite{wang2022ikeamanual}. VASR predicts the hierarchy from
part and target drawings without reading intermediate instruction
pages at inference. On IKEA, VASR scores 54.81\% on
Equivalent-Hard F1, compared with 32.40\% for Manual2Skill,
16.93\% for IKEA-GeoCluster, and 11.42\% for IKEA-SingleStep.
It also leads all three on the other four metrics. Compared with
Manual2Skill, the strongest task-specific baseline, VASR gains
22.41 percentage points (pp) on Hard F1 and 10.41~pp on ECR
(33.17\% versus 22.76\%).

On independent furniture, VASR outperforms Manual2Skill on all five
metrics. It improves Hard F1 from 29.89\% to 53.70\%
(23.81~pp) and ECR from 20.37\% to 31.74\% (11.37~pp).
We do not report SingleStep or GeoCluster on this set because it
lacks the 3D part point clouds for the IKEA-Manual evaluation
protocol. Thus, the task-specific comparison on independent
furniture is against Manual2Skill only.

\paragraph{Comparison with generalist VLMs.}
VASR also outperforms the strongest generalist baseline,
GPT6.0-Astra, on Equivalent-Hard F1 by 17.69~pp on IKEA
(54.81\% versus 37.12\%) and by 14.55~pp on independent furniture
(53.70\% versus 39.15\%). It ranks first among all compared methods
on all five metrics on IKEA and on four of five on independent
furniture. The exception is Equivalent-Simple F1 on the independent
set: Astra scores 68.71\%, ahead of VASR's 66.31\% by 2.40~pp.
Thus, Astra is better at recovering which parts belong to a
subassembly on this set, while VASR better recovers how those parts
are grouped and more often predicts a complete equivalent tree
(31.74\% versus 27.72\% ECR).

\paragraph{RL from the same supervised model.}
Standard GRPO lowers Equivalent-Hard F1 relative to VASR-S on both
datasets, despite small gains on Equivalent-Simple F1. TreeRPO and
Tree-GRPO improve both F1 metrics on both datasets, with Tree-GRPO the
strongest alternative RL method on Hard F1. Full VASR, trained with
RHPO, outperforms VASR-S on Hard F1 by 21.06~pp on IKEA and 20.88~pp
on independent furniture. Its gains over Tree-GRPO are 15.96~pp and
17.45~pp, respectively. The consistent ordering on the primary metric
supports the benefit of the full RHPO training procedure on both test
sets.

\subsection{Ablation Study}

\paragraph{Representation and planner components.}
We first isolate the contribution of the vector encoder and the
dedicated planning modules. Table~\ref{tab:vasr-component-ablation}
reports the results from the component ablations. Removing the SVG
encoder decreases Equivalent-Hard F1 from 54.81\% to 30.31\%, a
24.50~pp drop, indicating that native vector information provides
substantial evidence beyond the raster branch.
\begin{table}[!htbp]
  \centering
  \scriptsize
  \setlength{\tabcolsep}{2.5pt}
  \renewcommand{\arraystretch}{1.12}
  \resizebox{\columnwidth}{!}{%
  \begin{tabular}{lrrrrr}
    \toprule
    \textbf{Variant} &
    \textbf{Eq.-Simple} $\uparrow$ &
    \textbf{Eq.-Hard} $\uparrow$ &
    \textbf{ECS} $\uparrow$ &
    \textbf{ECH} $\uparrow$ &
    \textbf{ECR} $\uparrow$ \\
    \midrule
    VASR w/o SVG encoder
      & 48.82\% & 30.31\% & 50.14\% & 39.67\% & 21.72\% \\

    VASR-S
      & 50.43\% & 33.75\% & 56.06\% & 48.91\% & 24.49\% \\

    VASR w/o planner components
      & 38.12\% & 19.91\% & 36.32\% & 24.19\% & 11.76\% \\

    \rowcolor{vasrResultHighlight}
    \textbf{VASR (full)}
      & \vasrBest{67.73\%}
      & \vasrBest{54.81\%}
      & \vasrBest{77.35\%}
      & \vasrBest{62.43\%}
      & \vasrBest{33.17\%} \\
    \bottomrule
  \end{tabular}%
  }
  \caption{\textbf{Ablation of representation and planner components on
  the IKEA-derived test set.} All scores are percentages and report the
  mean over five random seeds.
  }
  \label{tab:vasr-component-ablation}
\end{table}

Replacing the dedicated
planner components reduces Equivalent-Hard F1 further to 19.91\%.
The supervised-only VASR-S obtains 33.75\% Equivalent-Hard F1, showing
that the post-supervised optimization stage is important for the full
system.

\begin{table*}[!htbp]
  \centering
  \small
  \setlength{\tabcolsep}{3.0pt}
  \renewcommand{\arraystretch}{1.15}
  \resizebox{\textwidth}{!}{%
  \begin{tabular}{lcccccc}
    \toprule
    \textbf{Variant} &
    \textbf{Repair-cost} &
    \textbf{Quality} &
    \textbf{Controlled} &
    \textbf{Action} &
    \textbf{Eq.-Hard F1} &
    \textbf{ECR} \\
    & \textbf{preference} &
    \textbf{gating} &
    \textbf{repair} &
    \textbf{distillation} &
    $\uparrow$ & $\uparrow$ \\
    \midrule
    Full RHPO & \cmark & \cmark & \cmark & \cmark
      & \vasrBest{54.81$\pm$0.72\%} & \vasrBest{33.17$\pm$0.85\%} \\
    w/o Repair Cost & \xmark & \cmark & \cmark & \cmark
      & 41.30$\pm$0.91\% & 27.80$\pm$1.12\% \\
    w/o Quality Gating & \cmark & \xmark & \cmark & \cmark
      & 48.24$\pm$1.05\% & 30.45$\pm$0.76\% \\
    w/o Controlled Repair & \cmark$^\star$ & \cmark & \xmark & \cmark$^\star$
      & 44.18$\pm$0.87\% & 28.62$\pm$0.45\% \\
    w/o Action Distillation & \cmark & \cmark & \cmark & \xmark
      & 46.72$\pm$0.54\% & 29.76$\pm$1.03\% \\
    \bottomrule
  \end{tabular}}
  \caption{\textbf{RHPO component ablation on the IKEA-derived test
  set.} Equivalent-Hard F1 and ECR are reported as percentages in
  mean$\pm$standard-deviation form over five random seeds. $\star$ marks the
  \textit{w/o Controlled Repair} setting: controlled repair is disabled
  during training, while repair cost is still measured at test time
  with a shared external evaluation protocol.}
  \label{tab:rhpo-ablation}
\end{table*}
\paragraph{RHPO component ablation.}
We next examine the contribution of the four central RHPO mechanisms:
repair-cost preference, quality gating, controlled repair, and
conservative action distillation. Table~\ref{tab:rhpo-ablation} reports
Equivalent-Hard F1 and ECR. Full RHPO obtains 54.81\% Equivalent-Hard
F1 and 33.17\% ECR. Removing repair-cost preference leads to the
largest decrease in Equivalent-Hard F1 among the listed single-component
ablations (41.30\%, a 13.51~pp drop) and lowers ECR to 27.80\%.
Removing controlled repair also causes a substantial decrease
(44.18\% Equivalent-Hard F1 and 28.62\% ECR). Removing action
distillation reduces the two metrics to 46.72\% and 29.76\%,
respectively. Without quality gating, Equivalent-Hard F1 and ECR fall
to 48.24\% and 30.45\%.

The ablation results indicate that repair-cost preference is a major
source of improvement: its removal reduces Equivalent-Hard F1 by
13.51~pp and ECR by 5.37~pp. Quality gating also contributes to the
final structure quality. In particular, removing the gate yields a
lower measured repair cost in the mechanism analysis, but this comes
with lower unassisted success and structural quality. This behavior
is consistent with the role of quality gating: repair cost should
distinguish candidates only after their unassisted outcomes are
sufficiently comparable. Controlled repair and action distillation
also provide complementary gains, as removing either component
decreases both complete-tree accuracy and node-level structural
agreement.

\paragraph{Mechanism analysis.}

To examine how the RHPO components affect decision quality and action
probabilities, we report the unassisted success estimate $Q$, structural
quality estimate $V$, expected future repair cost $C$, and the policy
probability mass assigned to verified-valid, verified-invalid, and
unknown actions. Let $P$, $N$, and $U$ denote the corresponding
verified-positive, verified-negative, and unknown action sets; their
masses are computed as
$M_P=\sum_{a\in P}\pi(a)$,
$M_N=\sum_{a\in N}\pi(a)$, and
$M_U=\sum_{a\in U}\pi(a)$.
Table~\ref{tab:rhpo-mechanism} reports the results.

The mechanism results provide evidence for the intended separation
between unassisted quality and repair burden. Removing repair-cost
preference leaves $Q$ and $V$ close to the full model (0.73 vs.\ 0.74
and 0.82 vs.\ 0.83), but increases $C$ from 1.20 to 1.78. This suggests
that repair cost contributes information not fully captured by
unassisted success and structural quality. Removing quality gating
reduces $C$ further to 0.95, but also lowers $Q$ from 0.74 to 0.64 and
$V$ from 0.83 to 0.75, illustrating the quality--repair trade-off that
the gate is designed to control. Finally, removing action distillation
reduces Valid Mass from 0.72 to 0.58 and increases Invalid Mass from
0.08 to 0.22, consistent with the intended redistribution of policy
probability away from verified-invalid actions and toward verified-valid
actions. The Unknown Mass remains close to 0.20 across variants,
suggesting that the method does not simply collapse probability onto
the currently verified action subset.
\begin{table}[!htbp]
  \centering
  \scriptsize
  \setlength{\tabcolsep}{2.0pt}
  \renewcommand{\arraystretch}{1.13}
  \resizebox{\columnwidth}{!}{%
  \begin{tabular}{lrrrrrr}
    \toprule
    \textbf{Variant} &
    \textbf{Success $Q$} $\uparrow$ &
    \textbf{Quality $V$} $\uparrow$ &
    \textbf{Repair Cost $C$} $\downarrow$ &
    \textbf{Valid Mass} $\uparrow$ &
    \textbf{Invalid Mass} $\downarrow$ &
    \textbf{Unknown Mass} \\
    \midrule
    \rowcolor{vasrResultHighlight}
    \textbf{Full RHPO} &
    \vasrBest{0.74$\pm$0.01} &
    \vasrBest{0.83$\pm$0.01} &
    \vasrBest{1.20$\pm$0.08} &
    \vasrBest{0.72$\pm$0.02} &
    \vasrBest{0.08$\pm$0.01} &
    0.20$\pm$0.01 \\

    w/o Repair Cost &
    0.73$\pm$0.01 & 0.82$\pm$0.02 & 1.78$\pm$0.10 &
    0.64$\pm$0.02 & 0.15$\pm$0.02 & 0.21$\pm$0.01 \\

    w/o Quality Gating &
    0.64$\pm$0.02 & 0.75$\pm$0.02 &
    \textbf{0.95$\pm$0.07} &
    0.59$\pm$0.03 & 0.20$\pm$0.02 & 0.21$\pm$0.01 \\

    w/o Controlled Repair &
    0.68$\pm$0.02 & 0.78$\pm$0.02 & 1.92$\pm$0.12 &
    0.63$\pm$0.02 & 0.16$\pm$0.02 & 0.21$\pm$0.01 \\

    w/o Action Distillation &
    0.70$\pm$0.02 & 0.80$\pm$0.01 & 1.50$\pm$0.09 &
    0.58$\pm$0.02 & 0.22$\pm$0.02 & 0.20$\pm$0.01 \\
    \bottomrule
  \end{tabular}%
  }
  \caption{\textbf{Mechanism analysis of RHPO.}
  $Q$ denotes unassisted success probability, $V$ denotes expected
  structural quality, and $C$ denotes the expected number of future
  repair interventions under the shared evaluation protocol.
  Valid, Invalid, and Unknown Mass are the policy probability masses
  assigned to verified-valid, verified-invalid, and unverified actions,
  respectively. Values are reported as mean$\pm$standard deviation
  over five random seeds. The lowest repair cost is boldfaced; the
  best values for the other directional metrics are highlighted in red.}
  \label{tab:rhpo-mechanism}
\end{table}

\section{Conclusion}
We presented VASR for predicting furniture assembly hierarchies from part and target illustrations.
The framework combines a dedicated SVG encoder with pretrained visual features, uses predicted connections to guide merge decisions, and accommodates admissible alternative structures through equivalence-aware supervision.
RHPO uses the cost of correcting verified errors in their subsequent rollouts to favor decisions when candidates achieve comparable unassisted quality.
VASR outperforms GPT6-Astra by 17.69 percentage points in Equivalent-Hard F1 on the IKEA-derived test set while maintaining strong performance on an independent furniture dataset.

\FloatBarrier
{%
  \small
  \raggedright
  \bibliographystyle{unsrtnat}
  \bibliography{main}

@inproceedings{wang2022ikeamanual,
  author = {R. Wang and Y. Zhang and J. Mao and R. Zhang and C.-Y. Cheng and J. Wu},
  title = {{IKEA-Manual}: Seeing Shape Assembly Step by Step},
  booktitle = {NeurIPS Datasets and Benchmarks},
  year = {2022}
}

@inproceedings{cao2023svgformer,
  author = {D. Cao and Z. Wang and J. Echevarria and Y. Liu},
  title = {{SVGformer}: Representation Learning for Continuous Vector Graphics Using Transformers},
  booktitle = {IEEE/CVF Conference on Computer Vision and Pattern Recognition},
  year = {2023},
  pages = {10093--10102}
}

@inproceedings{zou2024vgbench,
  author = {B. Zou and M. Cai and J. Zhang and Y. J. Lee},
  title = {{VGBench}: Evaluating Large Language Models on Vector Graphics Understanding and Generation},
  booktitle = {EMNLP},
  year = {2024},
  pages = {3647--3659}
}

@inproceedings{xing2025llm4svg,
  author = {X. Xing and J. Hu and G. Liang and J. Zhang and D. Xu and Q. Yu},
  title = {Empowering {LLMs} to Understand and Generate Complex Vector Graphics},
  booktitle = {IEEE/CVF Conference on Computer Vision and Pattern Recognition},
  year = {2025},
  pages = {19487--19497}
}

@article{goldberg2013oracles,
  author = {Y. Goldberg and J. Nivre},
  title = {Training Deterministic Parsers with Non-Deterministic Oracles},
  journal = {Transactions of the Association for Computational Linguistics},
  year = {2013},
  volume = {1},
  pages = {403--414}
}

@inproceedings{ross2011dagger,
  author = {S. Ross and G. Gordon and D. Bagnell},
  title = {A Reduction of Imitation Learning and Structured Prediction to No-Regret Online Learning},
  booktitle = {AISTATS},
  year = {2011},
  pages = {627--635}
}

@article{shao2024deepseekmath,
  author = {Z. Shao and others},
  title = {{DeepSeekMath}: Pushing the Limits of Mathematical Reasoning in Open Language Models},
  journal = {arXiv preprint arXiv:2402.03300},
  year = {2024}
}

@inproceedings{li2020singleimage,
  author = {Y. Li and K. Mo and L. Shao and M. Sung and L. Guibas},
  title = {Learning {3D} Part Assembly from a Single Image},
  booktitle = {ECCV},
  year = {2020},
  pages = {664--682}
}

@inproceedings{zhang2025manualpa,
  author = {J. Zhang and A. Cherian and C. Rodriguez and W. Deng and S. Gould},
  title = {{Manual-PA}: Learning {3D} Part Assembly from Instruction Diagrams},
  booktitle = {ICCV},
  year = {2025},
  pages = {6304--6314}
}

@inproceedings{guo2024treesba,
  author = {M. Guo and C. Li and Y. Zhao and G. H. Lee},
  title = {{TreeSBA}: Tree-Transformer for Self-Supervised Sequential Brick Assembly},
  booktitle = {ECCV},
  year = {2024}
}

@inproceedings{aslan2022assemblerl,
  author = {{\"O}. Aslan and B. Bolat and B. Bal and T. T{\"u}mer and E. {\c S}ahin and S. Kalkan},
  title = {{AssembleRL}: Learning to Assemble Furniture from Their Point Clouds},
  booktitle = {IROS},
  year = {2022},
  pages = {2748--2753}
}

@inproceedings{tie2025manual2skill,
  author = {C. Tie and others},
  title = {{Manual2Skill}: Learning to Read Manuals and Acquire Robotic Skills for Furniture Assembly Using Vision-Language Models},
  booktitle = {Robotics: Science and Systems},
  year = {2025}
}

@inproceedings{tie2026manual2skillplusplus,
  author = {C. Tie and others},
  title = {{Manual2Skill++}: Connector-Aware General Robotic Assembly from Instruction Manuals via Vision-Language Models},
  booktitle = {ICRA},
  year = {2026}
}

@article{qi2026assemstate,
  author = {Z. Qi and J. Li and Y. Shen and C. Qian and J. Liu},
  title = {{AssemState}: Manual and Physical-State-Guided Reasoning for Zero-shot Furniture Assembly},
  journal = {arXiv preprint arXiv:2610.08446v2},
  year = {2026}
}

@inproceedings{carlier2020deepsvg,
  author = {A. Carlier and M. Danelljan and A. Alahi and R. Timofte},
  title = {{DeepSVG}: A Hierarchical Generative Network for Vector Graphics Animation},
  booktitle = {NeurIPS},
  year = {2020}
}

@inproceedings{wang2026internsvg,
  author = {H. Wang and others},
  title = {{InternSVG}: Towards Unified {SVG} Tasks with Multimodal Large Language Models},
  booktitle = {ICLR},
  year = {2026}
}

@inproceedings{fan2022cadtransformer,
  author = {Z. Fan and T. Chen and P. Wang and Z. Wang},
  title = {{CADTransformer}: Panoptic Symbol Spotting Transformer for {CAD} Drawings},
  booktitle = {IEEE/CVF Conference on Computer Vision and Pattern Recognition},
  year = {2022},
  pages = {10986--10996}
}

@article{xing2026svgfusion,
  author = {X. Xing and J. Hu and Z. Xue and J. Zhang and B. Li and S. Wang and D. Xu and Q. Yu},
  title = {{SVGFusion}: A {VAE}-Diffusion Transformer for Vector Graphic Generation},
  journal = {arXiv preprint arXiv:2412.10437v3},
  year = {2026}
}

@article{ross2014aggrevate,
  author = {S. Ross and J. A. Bagnell},
  title = {Reinforcement and Imitation Learning via Interactive No-Regret Learning},
  journal = {arXiv preprint arXiv:1406.5979},
  year = {2014}
}

@inproceedings{chang2015lols,
  author = {K.-W. Chang and A. Krishnamurthy and A. Agarwal and Daum{\'e}, III, H. and J. Langford},
  title = {Learning to Search Better than Your Teacher},
  booktitle = {ICML},
  year = {2015},
  pages = {2058--2066}
}

@article{sutton1999temporal,
  author = {R. S. Sutton and D. Precup and S. Singh},
  title = {Between {MDPs} and {Semi-MDPs}: A Framework for Temporal Abstraction in Reinforcement Learning},
  journal = {Artificial Intelligence},
  year = {1999},
  volume = {112},
  number = {1--2},
  pages = {181--211}
}

@inproceedings{ji2026treegrpo,
  author = {Y. Ji and Z. Ma and Y. Wang and G. Chen and X. Chu and L. Wu},
  title = {Tree Search for {LLM} Agent Reinforcement Learning},
  booktitle = {ICLR},
  year = {2026}
}

@article{yang2025treerpo,
  author = {Z. Yang and Z. Guo and Y. Huang and X. Liang and Y. Wang and J. Tang},
  title = {{TreeRPO}: Tree Relative Policy Optimization},
  journal = {arXiv preprint arXiv:2506.05183v2},
  year = {2025}
}

@inproceedings{setlur2025rewarding,
  author = {A. Setlur and others},
  title = {Rewarding Progress: Scaling Automated Process Verifiers for {LLM} Reasoning},
  booktitle = {ICLR},
  year = {2025}
}

@inproceedings{li2022haco,
  author = {Q. Li and Z. Peng and B. Zhou},
  title = {Efficient Learning of Safe Driving Policy via Human-{AI} Copilot Optimization},
  booktitle = {ICLR},
  year = {2022}
}

@inproceedings{luo2024rlif,
  author = {J. Luo and P. Dong and Y. Zhai and Y. Ma and S. Levine},
  title = {{RLIF}: Interactive Imitation Learning as Reinforcement Learning},
  booktitle = {ICLR},
  year = {2024}
}

@inproceedings{huang2026findfixreason,
  author = {H. Huang and C. Qin and Y. Li and Y.-C. Chen},
  title = {Find, Fix, Reason: Context Repair for Video Reasoning},
  booktitle = {ICML},
  year = {2026},
  pages = {47455--47477}
}

@article{he2026pivotopd,
  author = {Y. He and others},
  title = {{PivotOPD}: Learning to Recover from Pivotal Mistakes in Multi-Turn Agents},
  journal = {arXiv preprint arXiv:2609.40285},
  year = {2026}
}

@article{chen2026mintrl,
  author = {M. Chen and Y. Tao and G. Friedland and X. Zhang and C. Kong},
  title = {{MInTRL}: Off-policy Intervention can boost On-policy {RL}},
  journal = {arXiv preprint arXiv:2609.12419},
  year = {2026}
}

@inproceedings{rusu2016distillation,
  author = {A. A. Rusu and others},
  title = {Policy Distillation},
  booktitle = {ICLR},
  year = {2016}
}

@article{bai2025qwen3vl,
  author = {S. Bai and others},
  title = {{Qwen3-VL} Technical Report},
  journal = {arXiv preprint arXiv:2511.21631},
  year = {2025}
}

@article{schulman2017ppo,
  author = {J. Schulman and F. Wolski and P. Dhariwal and A. Radford and O. Klimov},
  title = {Proximal Policy Optimization Algorithms},
  journal = {arXiv preprint arXiv:1707.06347},
  year = {2017}
}

@misc{zai2026glm53flash,
  author       = {{Z.ai}},
  title        = {{GLM-5.3-Flash}},
  howpublished = {Official model card},
  year         = {2026},
  url          = {https://huggingface.co/zai-org/GLM-5.3-Flash}
}

@misc{qwen2026qwen3827b,
  author       = {{Qwen Team}},
  title        = {{Qwen3.8-27B}},
  howpublished = {Official model card},
  year         = {2026},
  url          = {https://huggingface.co/Qwen/Qwen3.8-27B}
}

@misc{openai2026gpt6sol,
  author       = {{OpenAI}},
  title        = {{GPT-6 Astra System Card}: Appendix on {GPT-6 Sol} and {GPT-6 Luna}},
  howpublished = {System card appendix},
  month        = sep,
  year         = {2026},
  url          = {https://deploymentsafety.openai.com/gpt-6-astra/sec:appendix-sol-luna}
}

@misc{openai2026gpt6astra,
  author       = {{OpenAI}},
  title        = {{GPT-6 Astra} System Card},
  howpublished = {System card},
  month        = sep,
  year         = {2026},
  url          = {https://deploymentsafety.openai.com/gpt-6-astra}
}

@article{deepseekai2026deepseekv41flash,
  author  = {{DeepSeek-AI} and others},
  title   = {{DeepSeek-V4.1-Flash}: Pushing the Limits of {KV} Cache Compression},
  journal = {arXiv preprint arXiv:2609.19969},
  year    = {2026},
  url     = {https://arxiv.org/abs/2609.19969}
}
}

\end{document}